\documentclass[conference]{IEEEtran}
\IEEEoverridecommandlockouts
\usepackage{cite}
\usepackage{amsmath,amssymb,amsfonts}
\usepackage{algorithmic}
\usepackage{subfigure}
\usepackage{graphicx}
\usepackage{textcomp}
\usepackage{xcolor}
\def\BibTeX{{\rm B\kern-.05em{\sc i\kern-.025em b}\kern-.08em
    T\kern-.1667em\lower.7ex\hbox{E}\kern-.125emX}}
\begin{document}

\title{Performing Deep Recurrent Double Q-Learning for Atari Games\\{}
\thanks{978-1-7281-5666-8/19/\$31.00 ©2019 IEEE}
}

\author{\IEEEauthorblockN{Felipe Moreno-Vera}
\IEEEauthorblockA{\textit{Universidad Cat\'olica San Pablo} \\
Arequipa, Per\'u \\
felipe.moreno@ucsp.edu.pe}
}

\maketitle

\begin{abstract}
Currently, many applications in Machine Learning are based on defining new models to extract more information about data, In this case Deep Reinforcement Learning with the most common application in video games like Atari, Mario, and others causes an impact in how to computers can learning by himself with only information called rewards obtained from any action. There is a lot of algorithms modeled and implemented based on Deep Recurrent Q-Learning proposed by DeepMind used in AlphaZero and Go. In this document, we proposed deep recurrent double Q-learning that is an improvement of the algorithms Double Q-Learning algorithms and Recurrent Networks like LSTM and DRQN.
\end{abstract}


\section{Introduction}
Currently, there is an increase in the number in of applications in Reinforcement Learning. One recently application of Deep Reinforcement Learning (DRL) is self-driving cars \cite{leonsistema, Leon_2018}, another one is in games like AlphaZero (Go, Chess, etc) and video games like Mario, Top racer, Atari, etc. Deep Reinforcement Learning is considered as a third model in Machine Learning (with Supervised Learning and Unsupervised Learning) with a different learning model and architecture.

There are several methods of implementing these learning processes, where Q-Learning is a prominent algorithm, the Q value of a pair (state, action) contains the sum of all these possible rewards. The problem is that this sum could be infinite in case there is no terminal state to reach and, in addition, we may not want to give the same weight to immediate rewards as to future rewards, in which case use is made of what is called an accumulated reinforcement with discount: future rewards are multiplied by a factor $\gamma$ ∈ [0, 1] so that the higher this factor, the more influence future rewards have on the Q value of the pair analyzed.

\section{Background}

Sutton et al. \cite{drl_arch,drl_intro}  define various models to describe Reinforcement Learning and how to understand it. DeepMind was the first to achieve this Deep Learning with AlphaZero and Go game using Reinforcement Learning with Deep Q-Learning (DQN) \cite{dqn} and Deep Recurrent Q-Leaning (DRQN) \cite{drqn}, follow up by OpenAI who recently surpassed professional players in StarCraft 2 (Gramve created by Blizzard) and previously in Dota 2 developed by Valve. Chen et al. \cite{dqrnn} proposed a CNN based on DRQN using Recurrent Networks (a little variance of DRQN model using LSTM, the first neural network architecture that introduces the concept of memory cell \cite{lstm}, on agents actions to extract more information from frames.

\subsection{Deep Q-Learning (DQN)}

The first algorithm proposed by DeepMind was Deep Q-Learning, based on Q-Learning with experience replay \cite{dqn}, with this technique they save the last N experience tuples in replay memory. 
This approach is in some respects limited since the memory buffer does not differentiate important transitions and always overwrites with recent transitions due to the finite memory size N.

\begin{figure}[htbp]
\centering
  \includegraphics[width=9cm]{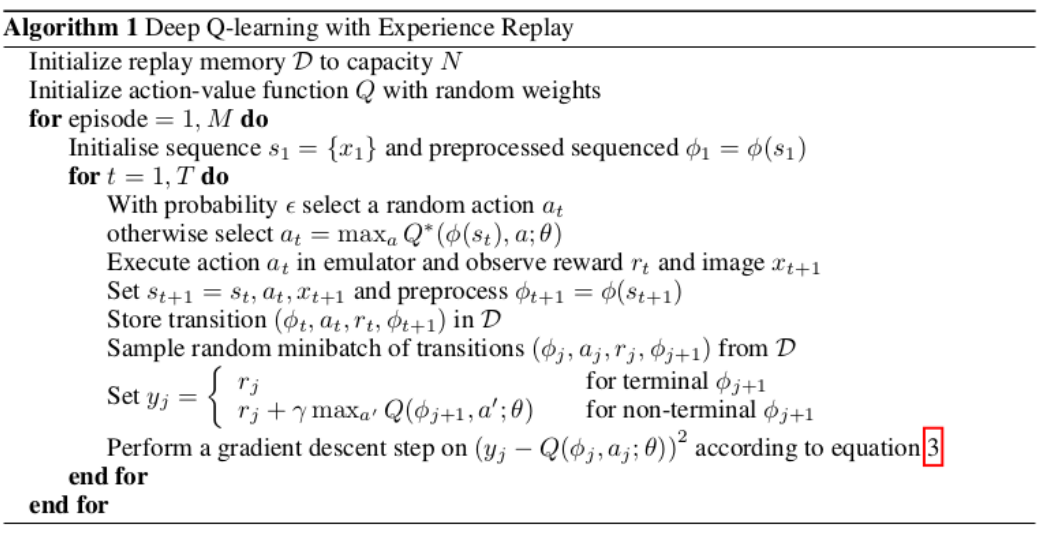}
  \caption{Deep Mind DQN algorithm with experience replay \cite{dqn}.}
  \label{fig:dqn}
\end{figure}

\subsection{Deep Double Q-Learning (DDQN)}

Hado et al. \cite{dqn} propose the idea of Double Q-learning is to reduce overestimation by decomposing the max operation in the target into action selection and action evaluation.

\begin{itemize}
    \item \textbf{DQN Model}: 
    $$
    Y_t = R_{t+1}+ \gamma max Q(S_{t+1}; a_t; \theta _t)
    $$
    \item \textbf{DDQN Model}:
    $$
    Y_t = R_{t+1}+ \gamma Q(S_{t+1}; argmax Q(S_{t+1}; a_t; \theta _t) ; \theta ^{1}_t)
    $$
\end{itemize}

Where:

\begin{itemize} \itemsep2pt
  \item $\bullet$ $a_t$ represents the agent.
  \item $\bullet$ $\theta_t$ are the parameters of the network.
  \item $\bullet$ Q is the vector of action values.
  \item $\bullet$ $Y_t$ is the target updated resembles stochastic gradient descent.
  \item $\bullet$ $\gamma$ is the discount factor that trades off the importance of immediate and later rewards.
  \item $\bullet$ $S_{t}$ is the vector of states.
  \item $\bullet$ $R_{t+1}$ is the reward obtained after each action.
\end{itemize}

\subsection{Deep Recurrent Q-Learning (DRQN)}

\begin{figure}[htbp]
\centering
  \includegraphics[width=7cm]{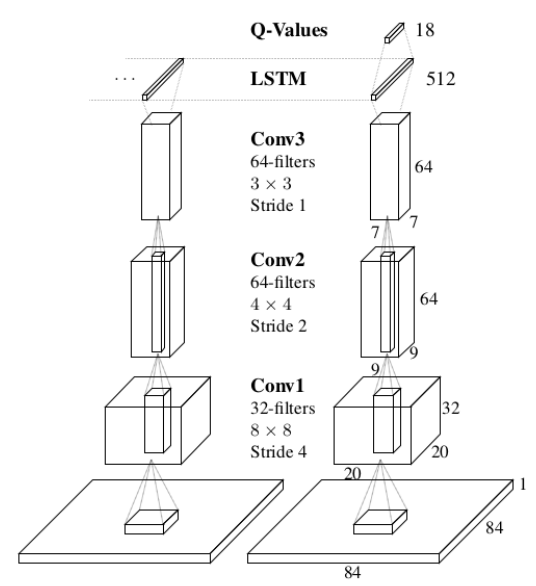}
  \caption{Deep Q-Learning with Recurrent Neural Networks model Deep Recurrent Q-Learning model (DQRN) \cite{drqn}}
  \label{fig:dqrn}
\end{figure}

Mathew et al. \cite{drqn} have been shown to be capable of learning human-level control policies on a variety of different Atari 2600 games. So they propose a DRQN algorithm which convolves three times over a single-channel image of the game screen. The resulting activation functions are processed through time by an LSTM layer (see Fig.\ref{fig:dqrn}.

\subsection{Deep Q-Learning with Recurrent Neural Networks (DQRNN)}

\begin{figure}[htbp]
\centering
  \includegraphics[width=7cm]{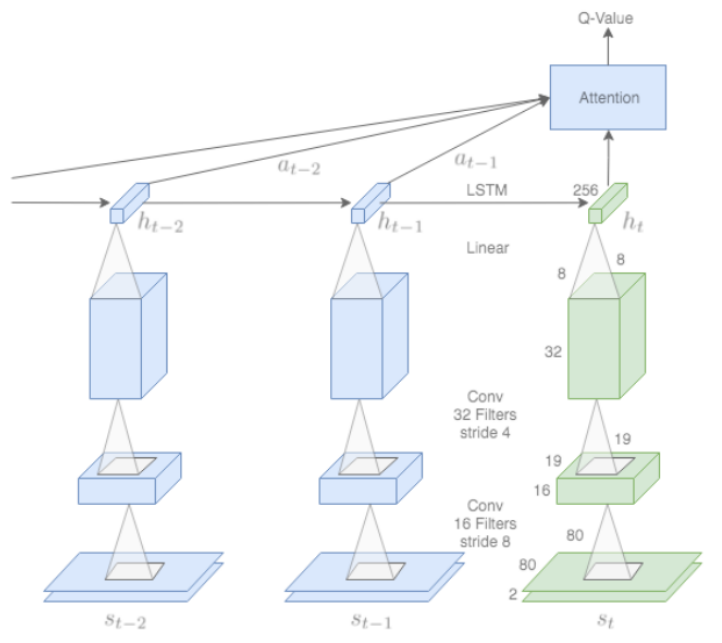}
  \caption{Deep Q-Learning with Recurrent Neural Networks model (DQRN) \cite{dqrnn}.}
  \label{fig:dqrnn}
\end{figure}

Chen et al. \cite{dqrnn} say DQN is limited, so they try to improve the behavior of the network using Recurrent networks (DRQN) using LSTM in the networks to take better advantage of the experience generated in each action (see Fig.\ref{fig:dqrnn}).

\section{Proposed model}

We implement the CNN proposed by Chen et al. \cite{dqrnn} with some variations in the last layers and using ADAM error. The first attempt was a simple CNN with 3 Conv 2D layers, with the Q-Learning algorithm, we obtain a slow learning process for easy games like SpaceInvaders or Pong and very low accuracy in complicated games like Beam Rider or Enduro. Then, we try modifying using Dense 512 and 128 networks at the last layer with linear activation and relu, adding an LSTM layer with activation tanh.

In table \ref{tab:hyperparameters} we present our Hyperparameters using in our models, we denote this list of hyperparameters as the better set (in our case). We run models over an NVIDIA GeForce GTX 950 with Memory 1954MiB using Tensorflow, Keras and GYM (Atari library) for python. We implement DDQN, DRQN, DQN and our proposed to combine DRQN with Double Q-Learning \cite{ddqn} algorithm using LSTM.

\begin{figure}[htbp]
\centering
  \includegraphics[width=9.5cm]{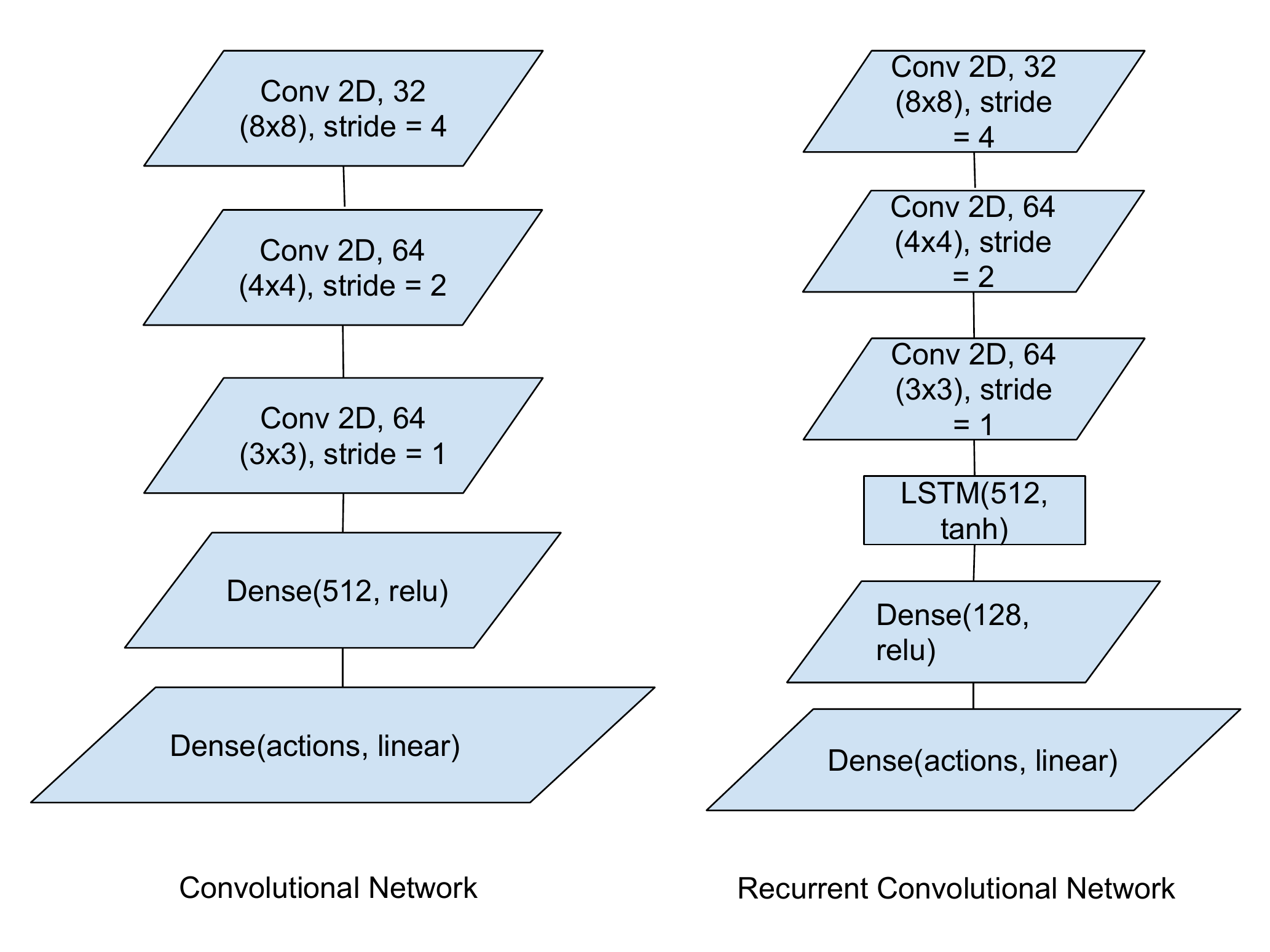}
  \caption{Convolutional Networks proposed in our models.}
  \label{fig:cnn}
\end{figure}

\section{Experiments and Results}

Our experiments are built over Atari Learning Environment (ALE) \cite{ale} which serves us as an evaluation platform for our algorithm in the games SpaceInvaders, Enduro, Beam Rider, and Pong and allow us to compare with DQN (Double Q-Learning), DDQN (Deep Double Q-Learning), and DRQN (Deep Recurrent Q-Learning). After to run our algorithms using 10M (10 million) episodes, we obtain results for each model in each respective game. We get the best scores for the 4 games mentioned above (See Table \ref{tab:scores}).

\begin{table}[htbp]
    \caption{Results Scores of Space Invaders, Enduro, Pong and Beam Rider.}
    \centering
      \begin{tabular}{|c|c|c|c|c|}
       \hline
        \multicolumn{5}{|c|}{\textbf{Models and respective Scores}} \\
        \hline
        Model & SpaceInvaders & Enduro & Pong & Beam Rider \\
        \hline
        DQN & 1450 & 1095 & 65 & 349\\
        \hline
        DRQN & 1680 & 885 & 39 & 594\\
        \hline
        DDQN & 2230 & 1283 & 44 & 167\\
        \hline
        DRDQN & 2450 & 1698 & 74 & 876\\
        \hline
      \end{tabular}
    \label{tab:scores}
\end{table}

\begin{table*}[!htb]
    \centering
    \caption{Hyperparameters used in models}
      \begin{tabular}{|l|c|l|}
       \hline
        \multicolumn{3}{|c|}{\textbf{List of Hyperparameters}} \\
        \hline
        Iterations & 10 000000 & number of batch iterations to the learning process \\
        \hline
        miniBatch size & 32 & number of experiences for SGD update\\
        \hline
        Memory buffer size & 900000 & SGD update are sampled from this number of most recent frames\\
        \hline
        Learning Rate & 0.00025 & learning rate used by RMS Propagation\\
        \hline
        Training Frequency & 4 & Repeat each action selected by the agent this many times \\
        \hline
        Y Update Frequency & 40000 & number of parameter updates after which the target network updates\\
        \hline
        Update Frequency & 10000 & number of actions by agent between successive SGD updates \\
        \hline
        Replay start size & 50000 & The number of Replay Memory in experience\\
        \hline
        Exploration max & 1.0 & Max value in exploration\\
        \hline
        Exploration min & 0.1 & Min value in exploration\\
        \hline
        Exploration Steps & 850000 & The number of frames over which the initial value of e reaches final value \\
        \hline
        Discount Factor & 0.99 & Discount factor $\gamma$ used in the Q-learning update\\
        \hline
      \end{tabular}
    \label{tab:hyperparameters}
\end{table*}

\begin{figure}[htbp]
\centering
  \includegraphics[width=8cm]{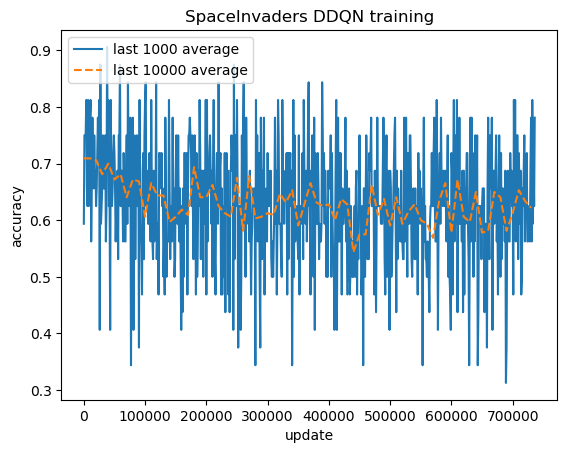}
  \caption{DDQN Accuracy.}
  \label{fig:ddqn_acc}
\end{figure}

\begin{figure}[htbp]
\centering
  \includegraphics[width=8cm]{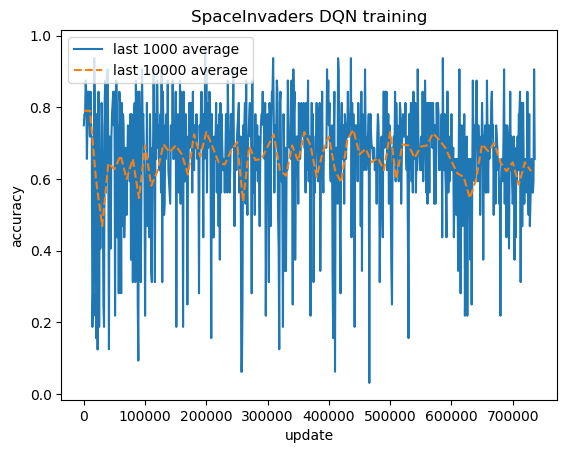}
  \caption{DRDQN Accuracy.}
  \label{fig:drdqn_acc}
\end{figure}

We compare with Volodymyr et al. \cite{drl} Letter about best scores form games obtained by DQN agents and professionals gamers (humans) to verify correct behavior of learning process, we measure accuracy based on Q-tables from the agent and DL algorithm (Double Q-Learning) extracting information from frames with the Convolutional Neural Networks (See Fig. \ref{fig:ddqn_acc}, and Fig.\ref{fig:drdqn_acc}).

\section*{Conclusions}
We present a model based on DRQN and Double Q-Learning combined to get a better performance in some games, using LSTM and CNN to analyze frames. We notice that each method could be good for a set of specific Atari games and other similar games but not for all. Some kind of sets of games can be improved using different CNN and get more information from the frames in each batch iteration.
In this work, we determine which one can be improved with the techniques of Deep Recurrent Double Q-Learning and which can be group based on which learning algorithm improve their scores.

\section*{Future Works}
With these results, we notice that every game in atari has a similar behavior to others different games, but because of technical details and our equipment, we could not determine which games haves similar behavior but we encourage to do it and get all set of similar behaviors and verify which methods should help to improve the learning process per each game.

\section*{Acknowledgment}
This work was supported by grant 234-2015-FONDECYT (Master Program) from CienciActiva of the National Council for Science,Technology and Technological Innovation (CONCYTEC-PERU).

\begin{thebibliography}{00}
\bibitem{Leon_2018}
Leon-Vera, Leonardo and Moreno-Vera, Felipe, Car Monitoring System in Apartments’ Garages by Small Autonomous Car Using Deep Learning, Annual International Symposium on Information Management and Big Data, Springer, 2018.

\bibitem{leonsistema}
Leon-Vera, Leonardo and Moreno-Vera, Felipe, Sistema de Monitoreo de Autos por Mini-Robot inteligente utilizando Tecnicas de Vision Computacional en Garaje Subterraneo, LACCEI, 2018.

\bibitem{drl_arch}
Richard S. Sutton, "Reinforcement Learning Architectures".

\bibitem{drl_intro}
 Richard Sutton and Andrew Barto. “Reinforcement Learning: An Introduction“. MIT Press, 1998.

\bibitem{dqn}
Volodymyr Mnih, Koray Kavukcuoglu, David Silver, Daan Wierstra, Alex Graves, Ioannis Antonoglou, Martin Riedmiller, "Playing Atari with Deep Reinforcement Learning", In NIPS Deep Learning Workshop 2013.

\bibitem{drqn}
Matthew Hausknecht and Peter Stone, "Deep Recurrent Q-Learning for Partially Observable MDPs", In AAAI Fall Symposium Series 2015.

\bibitem{dqrnn}
Clare Chen, Vincent Ying, Dillon Laird, "Deep Q-Learning with Recurrent Neural Networks".

\bibitem{lstm}
Hochreiter and Schmidhuber. “Long short-term memory“. Neural Comput. 9(8):1735-1780.

\bibitem{ddqn}
Hado van Hasselt, Arthur Guez and David Silver, "Deep Reinforcement Learning with Double Q-learning".

\bibitem{ale}
M. Bellemare, Y. Naddaf, J. Veness and M. Bowling, “The arcade learning enviroment. An evaluation platform for general agents“. In Journal of Artificial Intelligence Research, 47:253-279, 2013

\bibitem{drl}
Volodymyr Mnih, Koray Kavukcuoglu, David Silver, Andrei A. Rusu, Joel Veness, Marc G. Bellemare, Alex Graves, Martin Riedmiller, Andreas K. Fidjeland, Georg Ostrovski, Stig Petersen, Charles Beattie, Amir Sadik, Ioannis Antonoglou, Helen King, Dharshan Kumaran, Daan Wierstra, Shane Legg \& Demis Hassabis, "Human-level control through deep reinforcement learning".

\end{thebibliography}
\end{document}